\documentclass[12pt,a4paper,twoside]{article}
\usepackage[dvips]{epsfig}
\usepackage{graphicx}

\def\centerbmp#1#2#3{\vskip#2\relax\centerline{\hbox to#1{\special
  {bmp:#3 x=#1, y=#2}\hfil}}}

\usepackage[T1]{fontenc}

\usepackage{times}
\usepackage{fancyhdr}
\def\ps@firstpage{\let\@mkboth\@gobbletwo
\def\@oddhead{\sl TALN 2004, Session Poster, Fès, 19--21 avril 2004\hfil}
}

\def\@cite#1#2{(#1\if@tempswa ,
#2\fi)}
\def\@biblabel#1{}

\def\@citex[#1]#2{\if@filesw\immediate\write\@auxout{\string\citation{#2}}\fi
  \def\@citea{}\@cite{\@for\@citeb:=#2\do
    {\@citea\def\@citea{; }\@ifundefined
       {b@\@citeb}{{\bf ?}\@warning
       {Citation `\@citeb' on page \thepage \space undefined}}%
{\csname b@\@citeb\endcsname}}}{#1}}

\def\@internalcite\cite
\def\cite{\def\citename##1{##1}\@internalcite}
\def\shortcite{\catcode`:=12\def\citename##1{}\@internalcite}

\def\motsClefs#1#2{\section*{Mots-clefs -- Keywords}
#1\\
#2}
\makeatletter

\begin{document}

\title{An electronic dictionary as a basis for NLP tools: The Greek case}

\author{Ch. Tsalidis (1), A. Vagelatos (2) and G. Orphanos (1) \\
(1) Neurosoft S.A.\\
	   24 Kofidou Street\\
        GR-14231 Athens, Greece \\
(tsalidis,orphan)@neurosoft.gr \\
(2) R.A. Computer Technology Institute\\
	   13 Eptachalkou Street\\
        GR-11851 Athens, Greece \\
vagelat@cti.gr }

\date{}

\maketitle

\section*{R\'esum\'e - Abstract}

The existence of a Dictionary in electronic form for Modern Greek (MG) is mandatory if one is to process MG 
at the morphological and syntactic levels since MG is a highly inflectional language with marked stress and a 
spelling system with many characteristics carried over from Ancient Greek. Moreover, such a tool becomes necessary 
if one is to create efficient and sophisticated NLP applications with substantial linguistic backing and coverage. The present paper will focus on the deployment of such an electronic dictionary for Modern Greek, which was built in two phases: first it was constructed to be the basis for a spelling correction schema and then it was reconstructed in order to become the platform for the deployment of a wider spectrum 
of NLP tools.

\motsClefs{Lexique, morphologie}{Lexicon, morphology}
\newline
\newline
\newline
\newline
\newline
\newline

\pagebreak

\section{Introduction}

Electronic dictionaries have become among the most indispensable language resources for those involved in all aspects of natural language processing (NLP). Large--scale language resources (text and speech corpora, lexicons, grammars) are already developed or under development for an increasing number of natural languages.

Our computational linguistics (CL) team, over the past ten years has been conducting applied research toward the development of NLP applications and resources for Modern Greek (MG). 
The first step toward this goal was the design and development of a spelling corrector for Modern Greek. This corrector was based on a morphology lexicon. Later on, this morphology lexicon served as a basis for a number of research project as well as NLP application.

By the end of this project, we came to realise the need for a large-scale electronic dictionary, which could be the backbone for a wider and more advanced NLP systems as well as a valuable research tool. Such a dictionary should contain information for each entry (lemma) at the various linguistic levels: phonology, morphosyntax, syntax and semantics, as well as enable linking between the entries for various lexical and semantic relations: synonymy, antonymy, hyponymy, etc. 

In this paper we present a historical overview of the research activities of our CL team, regarding the development of an electronic dictionary. First the deployment of a morphology lexicon is described as well as various NLP tools that were based on it. Then the ``second phase'' of our research work is presented: the devise of a new coding scheme, the deployment of an electronic dictionary as well as some supporting tools and NLP applications. Finally we give our conclusions.

\section{How did we come here}

Back in 1991 our research unit at Computer Technology Institute (CTI - www.cti.gr) undertook a project to develop a spelling correction system for Modern Greek. That was the initiation for the foundation of a ``computational linguistics (CL) team'' that  started to develop a lexicon to be the basis of the whole project. 

The main goals for the lexicon construction (which were mostly focused on the spelling corrector that was the final target) were identified to be \cite{va:tr}: (a) {\it Storage economy}, (b) {\it Speed efficiency}, (c) {\it Dictionary coverage} and (d) {\it Optimum correction schema}. Under this framework, the linguistic analysis of MG led to the construction of a description language (which we called Greek Word Description Language - GWDL) that described both the inflectional morphology and marked stress of MG \cite{va:tr}. As a result, one and a half year later, a lexicon was developed that contained about 80,000 entries. The possible word forms produced from these entries have been calculated to exceed one million. In each entry, the stem(s) of a lexeme was(were) combined with the appropriate GWDL morphological rule(s), in order to produce all the corresponding word forms.

The primary storage mechanism used to access words in the Lexicon was the ``Compressed Trie'' \cite{KNU:aa}.  It was used as an index to the database of the words. This structure was relatively small (about 700Kb) compared to data needed to represent the entire lexicon. Thus, a big part of it (or the whole, if the computer had enough memory) could be loaded into main memory. The Compressed Trie contained the part of a word\rq s stem necessary to distinguish this word from stems of all other words having the same prefix \cite{va:tr}.

The correction schema was based on the well-known error categorization \cite{va:tr} of a) {\it orthographical errors} which are cognitive errors owing to the substitution of a deviant spelling for a correct one, when the author doesn\rq t know the correct spelling of a word or when he misconceives it and b) {\it typographical errors,} that are motoric errors, caused by hitting the wrong sequence of keys. 

Additionally, in MG we faced another error type, namely {\it stress position errors}, e.g. $``\kappa\acute\epsilon\phi\alpha\lambda\iota$'' /k$\epsilon$fali/ (head) instead of ``$\kappa\epsilon\phi\acute\alpha\lambda\iota$''.  The correction of this error type is based on the lexicon structure; words are stored in the lexicon without stress; stress is encoded in the rule part of each entry. Every word is searched without the stress; as soon as an entry has been matched, stress rules apply. If the stress is in a different position, then there is probably a stress position error and the word found is suggested as an alternative.

The lexicon was the heart of the spelling correction system (which later on was incorporated in the Greek version of Microsoft Office suite). Nevertheless, the lexicon did not serve only as a basis of this system but also in a number of other research applications like {\it stylistic analysis of poetic works} \cite{va:aa}, {\it computer assisted language learning (CALL)} \cite{sta:attt} and {\it word stemming} \cite{va:pe}.

\section{The new dictionary}

The importance of a more advanced and complete electronic dictionary, which would serve as a basis for a far more wide variety of NLP systems, became evident after the completion of the above mentioned ``first period'' of our NLP research team. At that time, it was decided to devote manpower as well as time in order to rebuild the existing lexicon with the following goals in mind: {\it the development of a new coding scheme able to incorporate appropriate annotation} (the information that has to be associated with each entry) and {\it the reconstruction of the lexicon\rq s data, based on a more appropriate (i.e. corpus based) methodology, taking into account a corpus that had been deployed for this purpose}.

Within this framework, the main requirements for the new dictionary were identified to be: (a) each lexicon entry should constitute a Lemma, (b) a lemma can contain one or more Lexemes (a lexeme is the representative of a cluster of morphological variations (word forms) of the same word), (c) all word forms of a lemma must be defined, (d) all word forms must be correctly hyphenated, (e) the various morphological parts of each word form (i.e. stem, suffix, prefix, etc.) must be identifiable, (f) stressing must be handled with an easy and efficient way, (g) a mechanism to incorporate simple (property) information as well as compound (structured) information should be supported, (h) full support of meanings of a lemma as in printed lexicography must be provided, (i) a power intra lemma reference mechanism should exist in order to represent ``wordnet'' links.

\subsection{Lexicon\rq s meta-language schema}

In order to fulfill the requirements of the lexicon, a coding scheme was devised to represent all this information and a special toll was implemented to permit the easy and efficient editing of lemma\rq s information. A lemma is defined as a set of lexemes: 

\begin{tabular}{ll}
 & lemma $\rightarrow$	name [lexeme]	\\
 & lexeme $\rightarrow$ name morphology semantics
\end{tabular}

\noindent where (in the formulas presented in this paper), [a] means one or more instances, \{a\} means zero or more instances, a $\mid$ b means a or b but not both, while a? means zero or one instance of a. As the above formula shows, a lexeme definition contains three parts. The {\it name} of the lexeme which identifies the lexeme, the {\it morphology} which describes how the lexeme\rq s word forms are constructed from their constituent parts and the {\it semantics} which holds the meanings information that can accompany a lemma.

The basic unit of word forms are the letters of MG alphabet. Despite this, words are usually divided in more complex parts called morphemes. Morphemes constitute the basic unit of word forms. We distinguish four types of morphemes: prefix, stem, infix and inflection (suffix). The first three types constitute the lexical-morpheme of the lexeme while the forth type is also known as the inflectional-morpheme. Formally, a lexeme\rq s morphology is defined as:
 
\begin{tabular}{ll}
morphology $\rightarrow$ lexical-morpheme, inflectional-morpheme stress \\
lexical-morpheme $\rightarrow$ [prefix $\mid$ stem $\mid$ infix] \\			
inflectional-morpheme $\rightarrow$ [inflection] \\
stress $\rightarrow$ [final $\mid$ penultimate $\mid$ antepenultimate] 
\end{tabular}

There are no stress characters inside the morphemes, while the position of the stress (final, penultimate, antepenultimate position) is attached as shown in the above formulas. Each morpheme must also be hyphenated so the derived word forms contain hyphenation information.

\subsection{Supporting tools}

In order to automate and simplify the coding of lexical information, various tools were constructed. One of them, LexEdit is a lightweight Lexicon Editor, which was used for the initial definition of the lexicon entries. Figure 1 shows a typical screen of LexEdit showing processed lexical entries. In the left pane of the application window we can see the sections that incorporate lemmas of the lexicon. We have a section for each MG alphabet letter. The ``iota'' $(\gamma\iota\acute\omega\tau\alpha - \iota)$ section is selected and in the right pane we have a part of the lemmas starting with the character ``iota''.

\begin{figure}[htbp]
\begin{center}
\scalebox{0.4}{\includegraphics{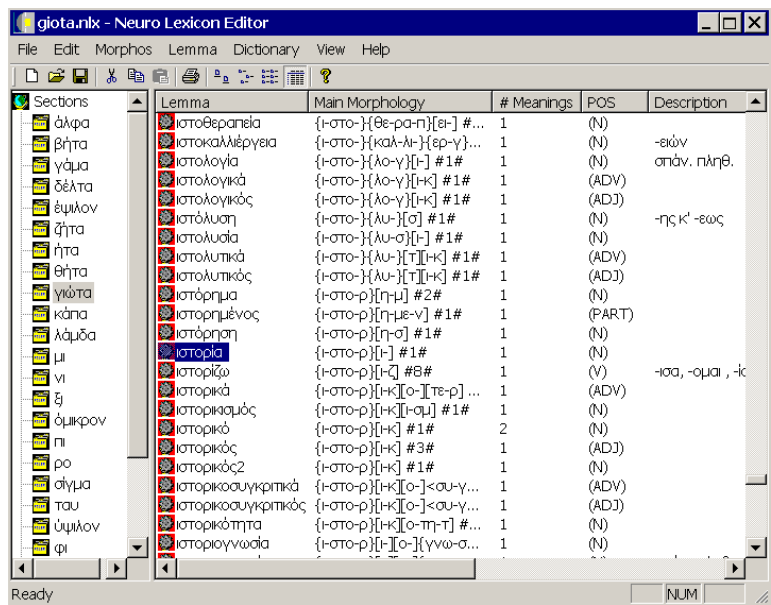}}
\caption{LexEdit tool}
\end{center}
\label{fig1}
\end{figure}

\subsection{The data}

The set of lemmas that were included in the lexicon was collected upon research based on the most important MG dictionaries (\cite{kti:a}, \cite{bab:in}, \cite{teg:fyt}) and on a corpus of MG texts.

The selection of lemmas is a particularly exigent and laborious work, which, with respect to MG, is also overloaded by the resent past of bilingualism (official - ``kathareyousa'' vs demotic - ``dimotiki'' languages). What is found in the dictionaries should be checked from different points of view. Since all the above dictionaries have not been based on text corpora, doubts arise for many entries, whether certain words or word forms or their meanings exist in MG (and not in dialects or ancient Greek).

\section{NLP Tools}

The new language tools that have been deployed are a result of systematic work of three and more years, at research level -in the areas of lexicography and NLP- as well as at the level of development of specialised electronic lexicons and computer systems for checking, correction and text hyphenation. Moreover, they are based on a redesign and reconstruction of the lexicon  as was described above. The tools (they can be found at the Web \cite{neu:so} are renewed two times per year. 

Thus far the following NLP tools have been implemented, based on the lexicon that was described above: a new spelling correction system and a hyphenator for MG.

The new spelling correction system includes approximately 90,000 lemmas (over 1,200,000 word forms). It was checked against a corpus of documents with more than 100,000,000 words. At the same time it includes approximately 200,000 English words thus it allows English and Greek spelling checking.

The search engine for the suggestions has been improved substantially. A number of new methods for correcting the errors have been added which are based on optical recognition (e.g., A - $\Delta$, T - $\Gamma$, $\Lambda\Lambda$ - M , $\alpha$ - $\sigma$), phonetic similarities (e.g., $\acute\alpha\nu\chi o\varsigma$ - $\acute\alpha\gamma\chi o\varsigma)$. Also, the methods for correcting the phonetic equivalences have been enhanced (e.g., $\acute\epsilon\beta\rho\epsilon\sigma\eta$ - $\epsilon\acute\upsilon\rho\epsilon\sigma\eta$) and have been enriched with methods to take care of the usual spelling errors like missing letter, added letter, transposed letter and wrong letter. Although the processing requirements have been increased, the new spelling engine is faster than the previous version.

The hyphenator  uses rules as well as the dictionary in order to achieve the best possible precision in hyphenation. 
Rules are separated in two categories: in those that were handcrafted according to the rules of hyphenation in MG grammar, and in those that were produced automatically based on hyphenation information incorporated in the lexicon. The rules of the second category enable the hyphenator to cope effectively with 26 vowel combinations, which in some words split during syllabification and in other not. 

Additionally, the verbal types that are liable to produce hyphenation errors as a result of the application of the hyphenation rules, have been incorporated in a list of exceptions. This list contains about 2.700 word forms containing vowel combinations, the syllabification of which leads to sense ambiguity.

\section{Conclusions}

The above presentation has hopefully succeeded in establishing an awareness of what is encompassed when refer to develop an electronic dictionary for Modern Greek. We have presented the two main development phases of this dictionary, we have cited limitations and we have presented numerous research as well as applied projects. More importantly, we have stressed and explained those features that characterize MG and which, in our point of view, make the dictionary in electronic form a necessary tool in all kinds of natural language processing.

\end{document}